\documentclass[letterpaper]{article} 
\usepackage{aaai25}  
\usepackage{times}  
\usepackage{helvet}  
\usepackage{courier}  
\usepackage[hyphens]{url}  
\usepackage{graphicx} 
\usepackage{natbib}  
\usepackage{caption} 
\usepackage{algorithm}
\usepackage{algorithmic}

\usepackage{tabularx}
\usepackage{cleveref}

\usepackage{xcolor}
\newcommand{\answerYes}[1]{\textcolor{blue}{#1}} 
\newcommand{\answerNo}[1]{\textcolor{teal}{#1}} 
\newcommand{\answerNA}[1]{\textcolor{gray}{#1}}

\usepackage{newfloat}
\usepackage{listings}
\DeclareCaptionStyle{ruled}{labelfont=normalfont,labelsep=colon,strut=off} 
\floatstyle{ruled}
\newfloat{listing}{tb}{lst}{}
\floatname{listing}{Listing}
\title{Leveraging Large Language Models for Automated Definition Extraction with TaxoMatic - a Case Study on Media Bias}

\author {
    Timo Spinde\textsuperscript{\rm 1},
    Luyang Lin\textsuperscript{\rm 2},
    Smi Hinterreiter\textsuperscript{\rm 3},
    Isao Echizen\textsuperscript{\rm 1}
}
\affiliations {
    \textsuperscript{\rm 1}National Institute of Informatics, Japan\\
    \textsuperscript{\rm 2}The Chinese University of Hong Kong, China\\
    \textsuperscript{\rm 3}University of Würzburg, Germany\\
    t.spinde@media-bias-research.org, lylin@se.cuhk.edu.hk, smi.hinterreiter@uni-wuerzburg.de, iechizen@nii.ac.jp
}

\usepackage{bibentry}

\begin{document}
\maketitle

\begin{abstract}
Defining complex, evolving concepts in academic research and extracting clear taxonomies from many publications is challenging. To streamline systematic reviews and capture shifts in conceptual understanding, we present our ongoing work on TaxoMatic - a framework leveraging Large Language Models (LLMs) to automate definition extraction from academic literature. The framework encompasses data collection, relevance classification to identify papers with definitions, and definition extraction using LLMs. As a first case study, we tested our relevancy evaluation component on 2,398 articles on media bias, a domain particularly rich in varying definitions and sub-concepts. Then, we evaluated our definition extraction component on manually reviewed papers, yielding 123 definitions from 113 relevant articles. Among five tested LLMs, Claude-3-sonnet achieved the highest F1 score (0.381) for relevance classification and demonstrated a median cosine similarity of 0.557 for definition extraction with role prompting. Future directions include improving relevance classification, expanding ground truth datasets, and applying this framework to other domains, potentially enhancing conceptual clarity across disciplines. 
\end{abstract}


 \begin{links}
     \link{Code/Dataset}{https://github.com/Media-Bias-Group/Taxomatic}
 \end{links}

\section{Introduction} \label{sec:intro}
Defining concepts and building taxonomies is a foundational research task, as it ensures methodological clarity and facilitates interdisciplinary communication \cite{spinde_media_2023}. However, extracting clear, systematic definitions from academic literature remains a significant challenge across domains, especially given the growing complexity of concepts and the rapid expansion of research output \cite{NEURIPS2023_abf3682c}.

Recent advances in large language models (LLMs) present new opportunities to streamline definition identification in academic research \cite{10.1007/978-3-031-70239-6_10}. Although LLMs have been explored for various information extraction and analysis tasks, their application to systematically extracting definitions and supporting conceptual clarity has not yet been largely investigated \cite{10.1007/978-3-031-70239-6_10}.

This study proposes and evaluates initial steps toward building TaxoMatic, a framework for automated definition extraction using LLMs. To assess its feasibility and reliability, we apply the individual parts of TaxoMatic to the domain of media bias, a concept widely studied in communication, political science, and computational linguistics but still lacking universally accepted definitions \cite{spinde_how_2022, wessel_introducing_2023, horych_promises_2024}. We address the following research questions:

\begin{itemize} 
\item \textbf{RQ1}: How accurately can LLMs evaluate the relevance of academic publications on media bias compared to human assessments?
\item \textbf{RQ2}: How do LLM and human-extracted definitions compare in content and semantic similarity?
\end{itemize}

We developed a three-stage workflow encompassing relevance classification, definition extraction, and evaluation. To create a ground truth, we collected 75,151 related scientific publications and manually rated 2,398 for relevancy to media bias research. From 113 deemed relevant, we manually extracted 123 definitions (\Cref{sec:methodology}).

The key contributions of this work include: 
\begin{itemize} 
\item Presenting an LLM-based process for systematic definition extraction from academic literature. 
\item Demonstrating the process's utility in extracting definitions of complex phenomena, focusing on media bias. 
\item Providing a dataset to evaluate future frameworks focusing on relevancy analysis and definition extraction. \end{itemize}

To ensure the dataset adheres to FAIR principles, we make all resources used in the process available (see the link after the abstract), use persistent identifiers for future updates, and maintain open, standardized formats to enhance interoperability and reusability. 

\section{Related Work}
\label{sec:related}
Definitions are the core of academic research, providing consistent communication and interpretation. When well established, researchers can engage in shared dialogue and consistently explore the same phenomenon \cite{navigli2010annotated}. Many domains, especially those studying human behavior, culture, or communication, face challenges in agreeing on unified definitions \cite{spinde_media_2023}. The phenomena they examine are often subjective, context-dependent, and influenced by multiple factors. For instance, in sociology and political science, concepts like democracy or social justice vary across cultural and ideological contexts. Similarly, media studies lack consensus on media bias \cite{spinde_interdisciplinary_2021}, with some focusing on visible bias like partisan reporting \cite{Groseclose2005}, while others define media bias as a linguistic concept \cite{spinde_neural_2021}. The definitional fragmentation complicates comparison across studies  \cite{spinde_media_2023}. Despite their importance, clear definitions are difficult to establish. Systematic reviews, which gather, analyze, and synthesize literature, often begin taxonomy-building but are labor-intensive and subjective. Researchers’ biases can influence the definitions, creating inconsistencies across studies \cite{krippendorff2019}. Additionally, manual analysis becomes impractical as datasets grow, slowing taxonomy development.

Traditional computational approaches for term extraction struggle with open-text documents due to their inability to effectively handle unstructured formats and context-dependent relationships \cite{bovi2015defie}. Advances in LLMs like GPT \cite{brown2020} and Mistral \cite{jiang2023}, built on transformer architectures, now enable automated definition extraction and support tasks like taxonomy building. LLMs excel at in-context learning, outperforming fine-tuned and unsupervised extractors in various domains, and their performance improves significantly with careful prompt engineering and experimenting with various prompt strategies \cite{10.1007/978-3-031-70239-6_10}. Still, their exact quality and reliability in extracting definitions and building taxonomies are yet unclear \cite{10.1007/978-3-031-70239-6_10}. Challenges persist, especially in domains with ambiguous or evolving terminologies and issues like hallucinations or dependency on predefined taxonomies \cite{10.1007/978-3-031-70239-6_10}.

Various datasets exist to explore definition extraction. The WCL dataset includes 5,000 sentences of explicit definitions from Wikipedia, limited to structured content \cite{navigli2010annotated}. The DEFT corpus, for SemEval-2010, covers 4,000 annotated sentences but lacks implicit or evolving definitions \cite{spala2020semeval}. DefIE contains 85,000 definitions from free-text sources across diverse domains \cite{bovi2015defie}. General dictionaries, like Oxford or Urban Dictionary\footnote{Which also exhibit inconsistent quality and informal style.}, collect various kinds of definitions \cite{oxford_dictionary, urban_dictionary}. Despite their value, all these datasets have limitations: (1) a lack of implicit or contested definitions (DEFT) \cite{spala2020semeval}, (2) focus on pre-structured content (WCL) \cite{navigli2010annotated}, (3) inconsistent quality (Urban Dictionary) \cite{urban_dictionary}, (4) limited adaptability to complex domains (all), and (5) no focus on academic definitions (all). These issues hinder their use for detailed, context-rich tasks, prompting the creation of our new dataset.

\section{Methodology}
\label{sec:methodology}
We show the general workflow of our TaxoMatic framework and the underlying case study in \Cref{tab:workflow}.
Our process includes multiple steps of a systematic literature search: (1) A keyword-based search for literature, (2) an assessment of which of the search results are relevant to the topic at hand, and (3) the extraction of required information --- in our case, definitions. Two intermediate steps are required to process this data with LLMs and to create our ground truth for evaluation. We describe all the steps in the following.

\begin{table}[h]
\caption{Workflow for the Definition Extraction Framework}
\label{tab:workflow}
\scriptsize 
\setlength{\tabcolsep}{1.5pt} 
\renewcommand{\arraystretch}{1} 
\begin{tabular}{|p{1.95cm}<{\raggedright}|p{1.95cm}<{\raggedright}|p{1.95cm}<{\raggedright}|p{1.95cm}<{\raggedright}|} 
\hline
\multicolumn{4}{|l|}{\textbf{Step 1: Searching Articles}} \\ \hline
1. Concepts list (4,200 keywords) & 2. Keyword review (1,096 keywords) & 3. Semantic Scholar: 578,447 papers & 4. 75,151 open-access PDFs (no duplicates) \\ \hline
\multicolumn{4}{|l|}{\textbf{Intermediate step: Data Preprocessing}} \\ \hline
\multicolumn{4}{|p{3.9cm}<{\raggedright}|}{1. PDF extraction via GROBID (63,038 XML files, ignoring files with errors)} \\ \hline
\multicolumn{4}{|l|}{\textbf{Intermediate step: Manual Ground Truth Preparation}} \\ \hline
\multicolumn{2}{|p{3.9cm}<{\raggedright}|}{1. 2,398 publications manually annotated for relevancy (Only articles with 100 or more citations)} & \multicolumn{2}{p{3.9cm}<{\raggedright}|}{2. 123 definitions manually extracted from 113 relevant articles} \\ \hline
\multicolumn{2}{|p{3.9cm}<{\raggedright}|}{\textbf{Step 2: Relevance Classification}} & \multicolumn{2}{p{3.9cm}<{\raggedright}|}{\textbf{Step 3: Definition Extraction}} \\ \hline
\multicolumn{2}{|p{3.9cm}<{\raggedright}|}{Automated labeling with 5 LLMs, 8 techniques (2,389 articles)} & \multicolumn{2}{p{3.9cm}<{\raggedright}|}{Automated labeling with Claude 3 Sonnet, 5 techniques (113 articles)} \\ \hline
\end{tabular}
\end{table}

\subsection{Dataset}
\label{sec:dataset}
\paragraph{Data Collection}
For our case study, we chose the media bias domain because of its large diversity and ambiguity of definitions \cite{spinde_media_2023, spinde_interdisciplinary_2021, spinde_automated_2021}, and collected data from Semantic Scholar using a keyword-based search.
As a seed for our search, we used the 21 terms from an existing but limited media bias taxonomy \cite{spinde_media_2023}. Aiming to cover as many media bias-related terms as possible, we expanded the list using GPT-3.5-turbo and generated 200 similar terms for each keyword\footnote{4,200 keywords in total, including duplicates.}. After removing duplicates, this resulted in a list of 1,096 unique keywords.
For each keyword, we crawled 1,000 results and eliminated duplicates and articles with fewer than 50 citations\footnote{We aim to add these in the future, but due to limited resources, initial filtering was required.}. Finally, we downloaded 75.151 open-access papers in PDF format.

\paragraph{Data Preprocessing}
To enable LLM processing of the PDF contents\footnote{We experimented with pasting entire PDFs, but most models do not allow it, and those that do showed poor performance. Using extracted text significantly improved results.}, we performed PDF information extraction with GROBID \cite{GROBID}. After manual verification and adjustments, we successfully processed $83.8\%$ of the PDFs (63,038 papers) into XML format for further use. 

\paragraph{Manual Definition Extraction}
To evaluate LLM performance, we manually annotated a ground truth. Due to limited reviewing capacity, we filtered the dataset by citation count \cite{bornmann2008citation}, selecting 2,398 articles with 100 or more citations. Six individuals, aged 25-35, with academic backgrounds and at least 6 months of media bias experience, followed two steps: reviewing titles and abstracts for relevancy and extracting or summarizing definitions from relevant full texts. Relevance was rated by one reviewer. Definition extraction involved a first reviewer and a second approving or modifying the result. This process produced 123 definitions from 113 relevant papers. While we believe the dataset size to be reasonable for our current evaluation, we aim to extend it, as discussed in \Cref{sec:discussion}.

\subsection{Publication Assessment \& Extraction}
\paragraph{LLM Selection} We selected five LLMs for the relevance classification task\footnote{We acknowledge the models change rapidly. We selected models based on the Hugginface Leaderboard and will update them in the future; see \Cref{sec:discussion}.}, GPT-3.5-turbo \cite{brown2020}, Mistral 7B Instruct v0.2 \cite{jiang2023}, Vicuna 13b v1.5 \cite{touvron2023},  Openchat 3.6 8b\cite{wang2023a}, and Claude 3 Sonnet \cite{anthropic2022}.

\paragraph{Prompting Strategies}
We designed our prompting strategies based on insights from prior research on prompt engineering, such as the importance of reasoning through Chain-of-Thought (CoT)\footnote{CoT prompting encourages the model to reason step-by-step, improving performance \cite{wei2022chain}.} prompting \cite{wei2022chain} and the effectiveness of giving contextually relevant examples \cite{zhou-etal-2023-context}. We used eight prompting strategies for the relevance classification, shown with examples in \Cref{appsec:prompting}. To select the four examples for the few-shot prompt, we applied two sampling strategies to ensure both relevance and diversity. First, for similarity sampling, we used the KATE (Knn-Augmented in-conText Example selection) strategy \cite{liu2021makes}, which identifies the most semantically or lexically similar examples based on Sentence-BERT (SBERT) embeddings \cite{reimers2019}. This ensured the selected examples closely matched the input context. Second, to enhance diversity, we applied k-means clustering to group the SBERT embeddings into clusters. Then, we manually selected two "relevant" and two "not relevant" examples from distinct clusters to capture a broader range of scenarios. For the definition extraction, we only use Claude 3 Sonnet with five of the eight strategies, namely Zero-Shot, Contextual Casual, Contextual Academic, CoT, and Role, since they focus on guiding the model's comprehension rather than relying on sampling strategies.

\paragraph{Experimental Setup}
We evaluated the relevance analysis with the 2,398 manually rated articles. Then, we analyzed the definition extraction using the 123 ground truth definitions from the 113 publications (see \Cref{sec:dataset}). Any assessment was run three times per model and prompting combination on Google Colab with the L4 GPU runtime using the Haystack library\footnote{See https://github.com/deepset-ai/haystack.}.

\subsection{Evaluation}
\paragraph{Relevance Classification}
We measured classification performance using Precision, Recall, F1-score, and Accuracy, and assessed label consistency with Krippendorff’s Alpha \cite{krippendorff2018content} per prompt.

\paragraph{Automated Definitions Extraction}
First, we used Sentence-BERT (SBERT) \cite{reimers2019} to embed LLM-extracted and manually extracted definitions and calculated their cosine similarity. Next, we applied a similarity threshold to classify matches, enabling the computation of precision, recall, and F1-score.

\section{Results}
\label{sec:results}

\subsection*{Relevance Classification}
As we show in \Cref{tab:relevancy_metrics}, Claude outperformed other models, achieving the highest F1-score (0.381) with balanced precision (0.440) and recall (0.482). OpenChat achieved the highest accuracy (0.803) but with less balanced precision (0.347) and recall (0.417), indicating a tendency to over-classify irrelevant items as relevant. Vicuna had the lowest accuracy (0.133), while Mistral recorded the lowest F1-score (0.100), making both unsuitable for relevance classification. ChatGPT performed average, with an F1-score of 0.216.

\begin{table}[h!]
    \centering
    \resizebox{\columnwidth}{!}{%
    \begin{tabular}{|l|c|c|c|c|}
        \hline
        \textbf{LLM} & \textbf{F1 Score} & \textbf{Accuracy} & \textbf{Precision} & \textbf{Recall} \\
        \hline
        ChatGPT-3.5 & 0.216 & 0.485 & 0.353 & 0.399 \\
        Mistral-7B & 0.100 & 0.200 & 0.283 & 0.302 \\
        OpenChat-3.6 & 0.339 & 0.803 & 0.347 & 0.417 \\
        Claude-3-sonnet & 0.381 & 0.672 & 0.440 & 0.482 \\
        Vicuna-13B & 0.102 & 0.133 & 0.377 & 0.379 \\
        \hline
    \end{tabular}
    }
    \caption{Average Relevance Classification Performance by Model}
    \label{tab:relevancy_metrics}
\end{table}

We show the results of different prompting strategies in \Cref{tab:prompting_metrics}. CoT performs best across all four metrics, followed by Role Prompting, demonstrating that step-by-step reasoning or adopting an expert role enhances performance. However, providing examples lowers performance across all metrics, especially with similar examples. Academic Contextual slightly outperforms Casual Contextual.

\begin{table}[h!]
    \centering
    \resizebox{\columnwidth}{!}{%
    \begin{tabular}{|l|c|c|c|c|}
        \hline
        \textbf{Prompting Strategy} & \textbf{F1 Score} & \textbf{Accuracy} & \textbf{Precision} & \textbf{Recall} \\
        \hline
        Zero-shot & 0.251 & 0.563 & 0.334 & 0.383 \\
        Contextual Similar Casual & 0.156 & 0.248 & 0.391 & 0.419 \\
        Contextual Similar Academic & 0.166 & 0.269 & 0.397 & 0.421 \\
        Contextual Diverse Casual & 0.158 & 0.403 & 0.266 & 0.307 \\
        Contextual Diverse Academic & 0.186 & 0.431 & 0.286 & 0.330 \\
        Chain-of-Thought (CoT) & 0.340 & 0.663 & 0.448 & 0.450 \\
        Role & 0.323 & 0.616 & 0.397 & 0.448 \\
        Emotional & 0.243 & 0.477 & 0.360 & 0.410 \\
        \hline
    \end{tabular}%
    }
    \caption{Average Relevance  Classification Performance by Strategy}
    \label{tab:prompting_metrics}
\end{table}

We find an overall Krippendorff’s Alpha of 0.162, suggesting some agreement. Across models, all five exhibit slightly negative values, indicating systematic disagreement based on the prompting technique. Among prompting techniques, CoT achieves the highest agreement (alpha = 0.702), showing strong model alignment with step-by-step reasoning. In contrast, Contextual Diverse Casual records the lowest agreement (alpha = 0.344), reflecting greater variability in classifications due to diverse examples.
More details of Krippendorff’s Alpha scores are shown in \Cref{appsec:alpha_results}.

\subsection*{Automated Definition Extraction}
In our automated definition extraction experiments, we exclusively used the Claude model, as it demonstrated the strongest performance in the Relevance Classification task. The cosine similarity scores for various prompting strategies are presented in \Cref{tab:cosine_similarity}.

Role Prompting achieved the highest mean similarity score (0.540), followed closely by Zero-shot Prompting (0.527). These findings indicate that the model's pre-trained knowledge was sufficient to grasp a broad understanding of media bias, even without additional contextual guidance. However, the relatively wide range of similarity scores reveals inconsistencies in the model's ability to capture slight details, particularly when distinguishing between explicit and implicit definitions.

\begin{table}[h!]
    \centering
    \resizebox{\columnwidth}{!}{%
    \begin{tabular}{|l|c|c|c|c|}
        \hline
        \textbf{Prompting Strategy} & \textbf{Mean} & \textbf{Median} & \textbf{Min} & \textbf{Max} \\
        \hline
        Zero-shot & 0.527 & 0.548 & 0.084 & 0.940 \\
        Contextual Casual & 0.508 & 0.525 & 0.138 & 0.895 \\
        Contextual Academic & 0.519 & 0.516 & 0.091 & 0.876 \\
        Chain-of-Thought (CoT) & 0.514 & 0.532 & 0.044 & 0.880 \\
        Role & 0.540 & 0.557 & 0.053 & 0.895 \\
        \hline
    \end{tabular}
    }
    \caption{Cosine Similarity Scores for Definition Extraction by Different Prompting Strategies}
    \label{tab:cosine_similarity}
\end{table}

To provide a more intuitive evaluation, we also applied a threshold-based approach using cosine similarity. Definitions with a similarity score above a 0.5 threshold were considered a correct match to the ground truth. This thresholding approach highlights Role Prompting's strengths, with the LLM achieving 70 correct definitions out of 113.

\begin{table}[h!]
    \centering
    \resizebox{\columnwidth}{!}{%
    \begin{tabular}{|l|c|c|c|}
        \hline
        \textbf{Prompting Strategy} & \textbf{Threshold 0.5} & \textbf{Threshold 0.6} & \textbf{Threshold 0.7} \\
        \hline
        Zero-shot & 62 & 44 & 25 \\
        Contextual Casual & 61 & 37 & 14 \\
        Contextual Academic & 59 & 36 & 19 \\
        CoT & 59 & 44 & 26 \\
        Role & 70 & 48 & 27 \\
        \hline
    \end{tabular}
    }
    \caption{Counts of Correct Definitions by Threshold Using Different Prompting Strategies}
    \label{tab:threshold_results}
\end{table}

\subsection*{Manual Error Analysis}

As part of the evaluation, we performed a manual qualitative review and identified several common model errors:

\paragraph{Overly Broad Definitions} In several cases, the LLM extracted overly general definitions that were semantically relevant but missed specific details of the bias discussed, particularly with zero-shot prompting.
\paragraph{Partial Definitions} Some definitions, especially with CoT Prompting, were incomplete, likely due to missing information during step-by-step reasoning.
\paragraph{Incorrect Definitions} Sometimes, the LLM extracted irrelevant or inaccurate text, misidentifying non-definitional content as definitions. This was observed more frequently in Contextual Casual Prompting, where the informal framing may have caused the model to focus on broader concepts.

\section{Discussion}\label{sec:discussion}

In this poster, we demonstrate the viability of using Large Language Models (LLMs) for extracting definitions from academic literature, with a case study on media bias. LLMs like Claude-3-sonnet effectively identify explicit and implicit definitions. Specifically addressing \textbf{RQ1}, Claude-3 showed high agreement with human assessments in relevance classification but struggled with class imbalance and overclassification. Regarding \textbf{RQ2}, LLMs aligned well with human definitions via SBERT cosine similarity, though subtle phrasing was often missed. TaxoMatic shows potential for broader applications, and our dataset offers a valuable resource for evaluating academic definition extraction.

Limitations include a class imbalance in relevance classification, overclassification, and the small ground truth dataset, which affects evaluation robustness. Cosine similarity, while useful, may miss slight phrasing differences.

Future work will expand the dataset with expert annotations and synthetic data to address class imbalance. Improved prompting strategies \cite{wei2023} and techniques like ontology learning and multi-stage modeling \cite{ji2021} could enhance performance. Regularly updating models and integrating processes into an automated framework could make TaxoMatic a versatile tool across disciplines.

\section{Conclusion}\label{sec:conclusion}

This poster introduced the initial steps and a dataset for TaxoMatic, an LLM-based framework for automating definition extraction from academic literature. Claude-3-sonnet achieved the best relevance classification performance while Chain-of-Thought and Role Prompting achieved the best definition extraction performance. Despite challenges like dataset imbalance and lexical variability, results show LLMs' potential to enhance conceptual clarity.

\section*{Acknowledgments}
The authors thank Diana Sharafeeva, Martin Spirit,  Fei Wu, Dr. Lingzhi Wang, and Prof. Dr. David Garcia. This work was supported by DAAD IFI, the JSPS KAKENHI Grants JP21H04907 and JP24H00732, by JST CREST Grants JPMJCR18A6 and JPMJCR20D3 including AIP challenge program, by JST AIP Acceleration Grant JPMJCR24U3, and by JST K Program Grant JPMJKP24C2 Japan.

{\fontsize{9pt}{11pt}\selectfont
\bibliography{refs}
}
\section*{Paper Checklist}

\appendix
\begin{enumerate}

\item For most authors...
\begin{enumerate}
    \item  Would answering this research question advance science without violating social contracts, such as violating privacy norms, perpetuating unfair profiling, exacerbating the socio-economic divide, or implying disrespect to societies or cultures?
    \answerYes{Yes, as it could facilitate finding and agreeing on definitions in different research domains.}
  \item Do your main claims in the abstract and introduction accurately reflect the paper's contributions and scope?
    \answerYes{Yes.}
   \item Do you clarify how the proposed methodological approach is appropriate for the claims made? 
    \answerYes{Yes, in \Cref{sec:methodology}.}
   \item Do you clarify what are possible artifacts in the data used, given population-specific distributions?
    \answerNA{NA. The study does not use population-specific data except for LLMs of Western origin.}
  \item Did you describe the limitations of your work?
    \answerYes{Yes, in \Cref{sec:discussion}.}
  \item Did you discuss any potential negative societal impacts of your work?
    \answerNA{The study evaluates LLM's ability to extract definitions. Hence, it could reinforce existing biases present in source materials, especially in domains with contested or politicized concepts. Additionally, reliance on LLMs may lead to overconfidence in automatically generated definitions.}
      \item Did you discuss any potential misuse of your work?
    \answerNo{No, besides the above mentioned bias and potential overconfidence.}
    \item Did you describe steps taken to prevent or mitigate potential negative outcomes of the research, such as data and model documentation, data anonymization, responsible release, access control, and the reproducibility of findings?
    \answerNA{NA. Code and data are publicly available. They do not contain personal data.}
  \item Have you read the ethics review guidelines and ensured that your paper conforms to them?
    \answerYes{Yes.}
\end{enumerate}

\item Additionally, if your study involves hypotheses testing...
\begin{enumerate}
  \item Did you clearly state the assumptions underlying all theoretical results?
    \answerYes{Yes, in \Cref{sec:methodology} and \Cref{sec:results}.}
  \item Have you provided justifications for all theoretical results?
    \answerYes{Yes, in \Cref{sec:results}.}
  \item Did you discuss competing hypotheses or theories that might challenge or complement your theoretical results?
    \answerNA{NA}
  \item Have you considered alternative mechanisms or explanations that might account for the same outcomes observed in your study?
    \answerNA{NA}
  \item Did you address potential biases or limitations in your theoretical framework?
    \answerYes{Yes, in \Cref{sec:discussion}.}
  \item Have you related your theoretical results to the existing literature in social science?
    \answerNA{NA}
  \item Did you discuss the implications of your theoretical results for policy, practice, or further research in the social science domain?
    \answerNA{NA}
\end{enumerate}

\item Additionally, if you are including theoretical proofs...
\begin{enumerate}
  \item Did you state the full set of assumptions of all theoretical results?
    \answerYes{Yes.}
	\item Did you include complete proofs of all theoretical results?
    \answerYes{Yes. All scripts and evaluations are available in the corresponding repository.}
\end{enumerate}

\item Additionally, if you ran machine learning experiments...
\begin{enumerate}
  \item Did you include the code, data, and instructions needed to reproduce the main experimental results (either in the supplemental material or as a URL)?
    \answerYes{Yes, under the repository URL.}
  \item Did you specify all the training details (e.g., data splits, hyperparameters, how they were chosen)?
    \answerYes{Yes.}
     \item Did you report error bars (e.g., with respect to the random seed after running experiments multiple times)?
    \answerNA{NA}
	\item Did you include the total amount of compute and the type of resources used (e.g., type of GPUs, internal cluster, or cloud provider)?
    \answerNA{NA}
     \item Do you justify how the proposed evaluation is sufficient and appropriate to the claims made? 
    \answerYes{Yes. See \Cref{sec:methodology} and \Cref{sec:results}.}
     \item Do you discuss what is ``the cost`` of misclassification and fault (in)tolerance?
    \answerNA{NA. We state the limitations of the definition extractions in \Cref{sec:discussion}.}
  
\end{enumerate}

\item Additionally, if you are using existing assets (e.g., code, data, models) or curating/releasing new assets, \textbf{without compromising anonymity}...
\begin{enumerate}
  \item If your work uses existing assets, did you cite the creators?
    \answerYes{Yes.}
  \item Did you mention the license of the assets?
    \answerYes{This project is licensed under the Apache 2.0 License. See the LICENSE file in the repository for details.}
  \item Did you include any new assets in the supplemental material or as a URL?
    \answerYes{Yes, under the repository URL.}
  \item Did you discuss whether and how consent was obtained from people whose data you're using/curating?
    \answerNA{NA. Data only contains academic open source paper.}
  \item Did you discuss whether the data you are using/curating contains personally identifiable information or offensive content?
    \answerNA{NA. Data does not contain personal data or offensive material.}
\item If you are curating or releasing new datasets, did you discuss how you intend to make your datasets FAIR?
\answerYes{Yes.}
\item If you are curating or releasing new datasets, did you create a Datasheet for the Dataset? 
\answerYes{Yes, in \Cref{sec:intro}.}
\end{enumerate}

\item Additionally, if you used crowdsourcing or conducted research with human subjects, \textbf{without compromising anonymity}...
\begin{enumerate}
  \item Did you include the full text of instructions given to participants and screenshots?
    \answerNA{NA}
  \item Did you describe any potential participant risks, with mentions of Institutional Review Board (IRB) approvals?
    \answerNA{NA}
  \item Did you include the estimated hourly wage paid to participants and the total amount spent on participant compensation?
    \answerNA{NA}
   \item Did you discuss how data is stored, shared, and deidentified?
   \answerNA{NA}
\end{enumerate}

\end{enumerate}

\section{Prompting Strategies}
\label{appsec:prompting}
\begin{enumerate}
    \item \textbf{Zero-Shot}: Minimal guidance without prior examples.
    \item \textbf{Contextual Similar Casual}: Similar examples in a casual context.
    \item \textbf{Contextual Similar Academic}: Similar examples in an academic context.
    \item \textbf{Contextual Diverse Casual}: Diverse examples in a casual context.
    \item \textbf{Contextual Diverse Academic}: Diverse examples in an academic context.
    \item \textbf{Chain-of-Thought Prompting (CoT)}: Step-by-step reasoning.
    \item \textbf{Role}: Model assumes the role of a media bias expert.
    \item \textbf{Emotional}: Uses emotional stimuli for deeper engagement.
\end{enumerate}

Examples for all of our prompts are published in the repository and below. 

\section*{Tested Prompts}
We used the following prompting strategies in our relevance assessment experiments. In each case, placeholders like \texttt{[Article Title]} and \texttt{[Article Abstract]} were replaced dynamically.

\begin{itemize}
    \item \textbf{Zero-Shot Prompting}\\
    \texttt{Please determine if the following article is relevant to media bias research: [Article Title] - [Article Abstract]}
    
    \item \textbf{Contextual Similar Casual}\\
    \texttt{Here are examples of articles relevant to media bias research: [Example 1], [Example 2]. Based on these, is the following article relevant? [Article Title] - [Article Abstract]}
    
    \item \textbf{Contextual Similar Academic}\\
    \texttt{Considering the provided scholarly articles on media bias: [Example 1], [Example 2], assess the relevance of this article to media bias research: [Article Title] - [Article Abstract]}
    
    \item \textbf{Contextual Diverse Casual}\\
    \texttt{We have diverse articles discussing various aspects of media studies: [Example 1], [Example 2]. Does the following article pertain to media bias? [Article Title] - [Article Abstract]}
    
    \item \textbf{Contextual Diverse Academic}\\
    \texttt{Given these diverse academic perspectives on media studies: [Example 1], [Example 2], evaluate if the following article is relevant to media bias research: [Article Title] - [Article Abstract]}
    
    \item \textbf{Chain-of-Thought (CoT) Prompting}\\
    \texttt{To determine if the following article is relevant to media bias research, let's analyze it step-by-step: [Article Title] - [Article Abstract]}
    
    \item \textbf{Role Prompting}\\
    \texttt{As a media bias expert, assess the relevance of this article to the field: [Article Title] - [Article Abstract]}
    
    \item \textbf{Emotional Prompting}\\
    \texttt{Imagine you're passionate about uncovering media bias. Does this article excite your interest in media bias research? [Article Title] - [Article Abstract]}
\end{itemize}

For the definition extraction, we used prompts as follows.

\begin{itemize}
    \item \textbf{Zero-Shot Prompting}\\
    \texttt{Extract the definition of media bias from the following academic text: [Full Text]}
    
    \item \textbf{Contextual Casual Prompting}\\
    \texttt{People often define media bias in different ways. Based on how it is discussed here, what is the definition? [Full Text]}
    
    \item \textbf{Contextual Academic Prompting}\\
    \texttt{In scholarly research, definitions are often embedded in complex texts. Please extract a clear, concise definition of media bias from the following excerpt: [Full Text]}
    
    \item \textbf{Chain-of-Thought (CoT) Prompting}\\
    \texttt{Let's identify the definition of media bias step by step. First, find any sentence that discusses the nature of media bias. Then, summarize that into a clear definition. Here is the article content: [Full Text]}
    
    \item \textbf{Role Prompting}\\
    \texttt{You are a researcher in media studies. Based on the following academic text, please provide the clearest definition of media bias presented in the article: [Full Text]}
\end{itemize}

\newpage
\section{Details of Krippendorff’s Alpha Results}
\label{appsec:alpha_results}

\begin{table}[h!]
    \centering
    \resizebox{\columnwidth}{!}{%
    \begin{tabular}{|l|c|}
        \hline
        \textbf{LLM} &  \textbf{Krippendorff’s Alpha} \\
        \hline
        ChatGPT-3.5 & -0.027  \\
        Mistral-7B & -0.054  \\
        OpenChat-3.6 & -0.032 \\
        Claude-3-sonnet & -0.063  \\
        Vicuna-13B & -0.108 \\
        \hline
    \end{tabular}
    }
    \caption{Krippendorff’s Alpha per LLM across all prompting techniques}
    \label{tab:alpha_per_llm}
\end{table}

\begin{table}[ht]
    \centering
    \resizebox{\columnwidth}{!}{%
    \begin{tabular}{|l|c|}
        \hline
        \textbf{Prompting Strategy} &  \textbf{Krippendorff’s Alpha} \\
        \hline
        Zero-shot &  0.610\\
        Contextual Similar Casual & 0.575 \\
        Contextual Similar Academic & 0.581 \\
        Contextual Diverse Casual & 0.344 \\
        Contextual Diverse Academic & 0.491 \\
        Chain-of-Thought (CoT) & 0.702 \\
        Role & 0.586 \\
        Emotional & 0.512 \\
        \hline
    \end{tabular}
    }
    \caption{Krippendorff’s Alpha per Prompting Technique across all Models}
    \label{tab:alpha_per_promptingstrategy}
\end{table}

\end{document}